\documentclass{bmvc2k}

\title{Deep Learning for Detecting Multiple Space-Time Action Tubes in Videos}

\addauthor{Suman Saha}{suman.saha-2014@brookes.ac.uk}{1}
\addauthor{Gurkirt Singh}{gurkirt.singh-2015@brookes.ac.uk}{1}
\addauthor{Michael Sapienza}{michael.sapienza@eng.ox.ac.uk}{2}
\addauthor{Philip H. S. Torr}{philip.torr@eng.ox.ac.uk}{2}
\addauthor{Fabio Cuzzolin}{fabio.cuzzolin@brookes.ac.uk}{1}

\addinstitution{
Dept. of Computing and Communication Technologies\\
Oxford Brookes University\\
Oxford, UK
}

\addinstitution{
Department of Engineering Science\\
University of Oxford\\
Oxford, UK
}

\runninghead{Saha \bmvaEtAl}{Deep Learning for detecting Space-Time Action Tubes}
\def\etal{\emph{et al}\bmvaOneDot}

\usepackage{graphicx}
\usepackage{epsfig}
\usepackage{booktabs}
\usepackage{cuted}
\usepackage{latex_pkgs/maths}
\usepackage{latex_pkgs/msformatting}
\usepackage{algorithm}
\usepackage{algorithmic}
\newcommand{\FuncProg}[1]{\text{\tt{#1}}} % function

\newcommand{\MIKE}[1]{MIKE: \textcolor{red}{#1}}

\begin{document}

\maketitle

%%%%%%%%% ABSTRACT
\begin{abstract}  
In this work, we propose an approach to the spatiotemporal localisation (detection) and classification of multiple concurrent actions within temporally untrimmed videos.
Our framework is composed of three stages.
In stage~1, appearance and motion detection networks are employed to localise and score actions from colour images and optical flow. 
In stage~2,
the appearance network detections are boosted by combining them with the motion detection scores, in proportion to their respective spatial overlap.
In stage~3, sequences of detection boxes most likely to be associated with a single action instance, called {action tubes},
are constructed by solving two energy maximisation problems via dynamic programming.
While in the first pass,
action paths spanning the whole video are built by linking detection boxes over time using their class-specific scores and their spatial overlap,
in the second pass, temporal trimming is performed by ensuring label consistency for all constituting detection boxes.
We demonstrate the performance of our algorithm on the challenging UCF101, J-HMDB-21 and LIRIS-HARL datasets,
achieving new state-of-the-art results across the board
and significantly increasing detection speed at test time.

\end{abstract}

%%%%%%%%% BODY TEXT
\section{Introduction} \label{sec:intro}

Recent advances in object detection via convolutional neural networks (CNNs)~\cite{girshick-2014} have triggered a significant 
performance improvement in the state-of-the-art action detectors~\cite{Georgia-2015a,Weinzaepfel-2015}. 
However, the accuracy of these approaches is limited by their relying on unsupervised region proposal 
algorithms such as Selective Search~\cite{Georgia-2015a} or EdgeBoxes~\cite{Weinzaepfel-2015} which, besides being resource-demanding,
cannot be trained for a specific detection task and are disconnected from the overall classification objective.
% Fabio: I thought the previous version read nicer while occupying pretty much the same space (below)
% TODO - check this with Fabio.
%Moreover, in \cite{Georgia-2015a}, the proposal classification stage is also disjoint from the learning of action features and proposals
%as they employ one-vs-all SVMs on the output of the CNN features.
%We found in practice that such disjoint action detection strategies which involve CNN fine-tuning, intensive feature extraction, storage,
%and classification give modest performance and are very computationally expensive.
Moreover, these approaches % Question: does this apply only to Georgia-2015a or also to Weinzaepfel-2015?
are computationally expensive as they follow a multi-stage
classification strategy which requires CNN fine-tuning and intensive feature extraction (at both
training and test time), the caching of these features onto disk, and finally the training of
a battery of one-vs-all SVMs for action classification.
On large datasets such as UCF-101~\cite{soomro-2012}, overall training and feature extraction takes a week using 7 Nvidia Titan X GPUs, %\footnote{.},
plus one extra day for SVM training.
At test time, detection is slow as features need to be extracted for each region proposal via a CNN forward pass.

To overcome these issues we propose a novel action detection framework which,
%overcomes the aforementioned drawbacks of the state-of-the-art action detectors.
instead of %solving the action classification and localisation tasks using 
adopting an expensive multi-stage pipeline,
takes advantage of the most recent single-stage deep learning architectures for object detection~\cite{ren2015faster},
in which a single CNN is trained for both detecting and classifying frame-level region proposals in an end-to-end fashion.
Detected frame-level proposals are subsequently linked in time to form space-time `action tubes'\cite{Georgia-2015a} by solving two optimisation problems via dynamic programming.
%By saying a single-stage training strategy we refer: training a single CNN for classification as compared to the multi-stage training approach 
%which requires both CNN and SVM training.
%using a pair of end-to-end trainable CNNs and a \emph{two-pass dynamic programming} algorithm.
%reducing space-time complexity of both training and testing, while significantly improving detection accuracy.
%--- REVIEWER-1: The aggressive claim over computational expensive is not justified. ---
We demonstrate that the proposed action detection pipeline is at least $2\times$ faster in training
and $5\times$ faster in test time detection speeds as compared to~\cite{Georgia-2015a,Weinzaepfel-2015}.
In the supplementary material, we present a comparative analysis of the training and testing time requirements of our approach with respect to~\cite{Georgia-2015a,Weinzaepfel-2015}
on the UCF-101~\cite{soomro-2012} and J-HMDB-21~\cite{J-HMDB-Jhuang-2013} datasets.
Moreover, our pipeline consistently outperforms previous state-of-the-art results (\S~\ref{subsec:discussion}).
\iffalse
across the challenging UCF-101,
J-HMDB-21 and LIRIS-HARL datasets (e.g. with an mAP of \textbf{66.75}\% vs. 46.77\% from~\cite{Weinzaepfel-2015} on UCF-101).
\fi
%,and is able to process 5 frames per second at test time.
% Fabio: I think relative speed improvement is not quite stressed enough..

\begin{figure}[tb]
  \centering
\includegraphics[width=0.98\textwidth]{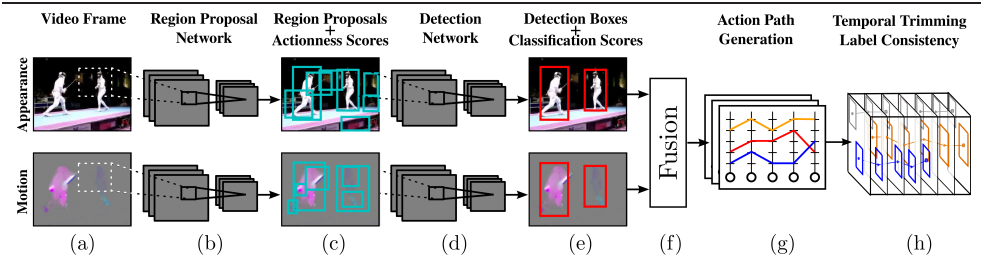}
  \vskip -0.2cm
  \caption{
    {\small
      \textit{
    At test time,
    \textbf{(a)}~RGB and optical-flow images are passed to
    \textbf{(b)}~two separate region proposal networks (RPNs).
    \textbf{(c)}~Each network outputs region proposals with associated actionness scores (\S~\ref{subsec:rpn}).
    \textbf{(d)}~Each appearance/motion detection network takes as input the relevant image and RPN-generated region proposals, and
    \textbf{(e)}~outputs detection boxes and softmax probability scores (\S~\ref{subsec:frcnn}).
    \textbf{(f)}~Appearance and motion based detections are fused (\S~\ref{subsec:fs_spatil_flow}) and
    \textbf{(g)}~linked up to generate class-specific action paths spanning the whole video.
    \textbf{(h)}~Finally the action paths are temporally trimmed to form action tubes (\S~\ref{subsec:action_tube}).
           }
    }
 }
  \label{fig:algorithmOverview} \vspace{-2mm}
\end{figure}

% stages of the methodology
\textbf{Overview of the approach}. 
Our approach is summarised in Fig.~\ref{fig:algorithmOverview}. 
We train two pairs of Region Proposal Networks (RPN)~\cite{ren2015faster} and Fast R-CNN~\cite{girshick2015fast} detection networks
- one on RGB and another on optical-flow images~\cite{Georgia-2015a}.
%Training two separate CNNs on spatial and flow features is a commonly employed strategy to boost action detection accuracy~\cite{Simonyan-2014,Georgia-2015a}.
For each pipeline, the RPN \textbf{(b)}, takes as input a video frame \textbf{(a)},
and generates a set of region proposals \textbf{(c)},
%\footnote{Rectangular region hypotheses in a video frame likely to contain actions of interest.}
and their associated `actionness' \cite{chen2014actionness} scores\footnote{A softmax score for a region proposal containing an action or not.
}.
% (b) Fast RCNN
Next, a Fast R-CNN~\cite{ren2015faster} detection network \textbf{(d)} takes as input the original video frame and a subset of the region proposals generated by the RPN,
and outputs a `regressed' detection box and a softmax classification score for each input proposal,
indicating the probability of an action class being present within the box.
% (c) merging of cues
To merge appearance and motion cues,
%rather than concatenating the intermediate layer (fc7) features from the two CNNs before classification~\cite{Simonyan-2014,Georgia-2015a},
we fuse \textbf{(f)} the softmax scores from the appearance- and motion-based detection boxes \textbf{(e)} (\S~\ref{subsec:fs_spatil_flow}).
We found that this strategy significantly boosts detection accuracy.

\begin{figure}[t]
\centering
\includegraphics[width=0.96\textwidth]{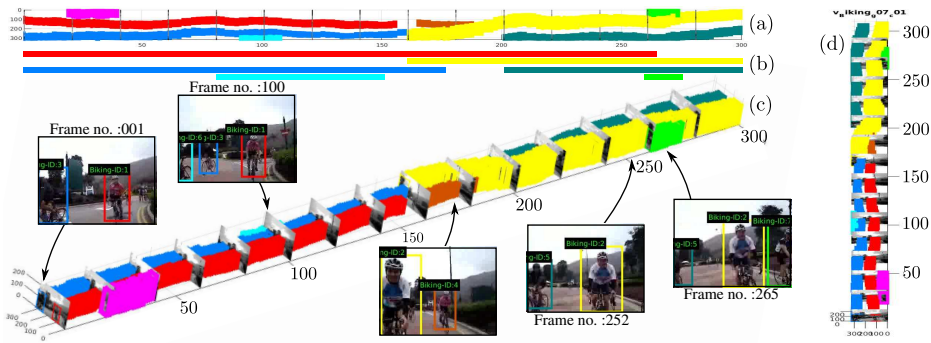}
\vskip -.3cm
\caption{
  {\small
  \textit{
Action tube detection in a `biking' video taken from UCF-101~\cite{soomro-2012}.
\textbf{(a)} Side view of the detected action tubes where each colour represents a particular instance.
The detection boxes in each frame are linked up to form space-time action tubes.
\textbf{(b)} Illustration of the ground-truth temporal duration for comparison.
\textbf{(c)} Viewing the video as a 3D volume with selected image frames;
notice that we are able to detect multiple action instances in both space and time.
\textbf{(d)} Top-down view.
  }
  }
}
  \label{fig:introductionTeaser} \vspace{-5mm}
\end{figure}

% (d,e) action path and tubes
After fusing the set of detections over the entire video,
we identify sequences of frame regions most likely to be associated with a single action tube. %  instance (action tubes). 
Detection boxes in a tube need to display a high score for the considered action class,
as well as a significant spatial overlap for consecutive detections.
Class-specific action paths \textbf{(g)} spanning the whole video duration are generated via a {Viterbi} forward-backward pass (as  in~\cite{Georgia-2015a}). An additional second pass of dynamic programming is introduced to take care of temporal detection \textbf{(h)}. 
As a result, our action tubes are not constrained to span the entire video duration, as in~\cite{Georgia-2015a}.
Furthermore, extracting multiple paths allows our algorithm to account for multiple co-occurring instances of the same action class (see Fig.~\ref{fig:introductionTeaser}).

%--- REVIEWER-1:    The only novelty is the generation of action tubes from the detected and labelled bounding boxes. This need to be emphasized in the article.---
% novelty-1
%\textcolor{red}{
Although it makes use of existing RPN~\cite{ren2015faster} and Fast R-CNN~\cite{girshick2015fast} architectures,
this work proposes a radically new approach to spatiotemporal action detection which 
brings them together with a novel late fusion approach and an original action tube
generation mechanism to dramatically improve accuracy and detection speed.
% novelty-2
Unlike~\cite{Georgia-2015a,Weinzaepfel-2015}, in which appearance and motion information are fused by combining fc7 features,
we follow a late fusion approach~\cite{Simonyan-2014}.
Our novel fusion strategy boosts the confidence scores of the detection boxes based on their spatial overlaps and their
class-specific softmax scores obtained from appearance and motion based networks~(\S~\ref{subsec:fs_spatil_flow}).
The 2$^{\textnormal{nd}}$ pass of dynamic programming, we introduce for action tube temporal trimming,
contributes to a great extent to significantly improve the detection performance~(\S~\ref{subsec:discussion}).

\iffalse
    %TODO: put this in another figure, or in video.
    Note that no action is detected at the beginning of the video although 
    (irrelevant) human motion is present there.
\fi

\textbf{Contributions.} In summary, 
this work's main contribution is a novel action detection pipeline which:

\begin{itemize}[leftmargin=0.5cm,noitemsep,nolistsep]

\item incorporates
 % the most
  recent deep Convolutional Neural Network architectures for simultaneously predicting frame-level detection boxes and the associated action class scores (\S~\ref{subsec:rpn}-\ref{subsec:frcnn});

\item uses an original fusion strategy for merging appearance and motion cues based on the softmax probability scores and spatial overlaps of the detection bounding boxes (\S~\ref{subsec:fs_spatil_flow});

\item brings forward a two-pass dynamic programming (DP) approach for constructing space time action tubes (\S~\ref{subsec:action_tube}).

\iffalse
\item a comparison between the quality of RPN and Selective Search based region proposals for human action detection on the UCF101 dataset (\S~\ref{sec:exp-proposals});
\fi

\end{itemize}
\noindent
An extensive evaluation on the main action detection datasets demonstrates that our approach significantly outperforms the current state-of-the-art,
and is 5 to 10 times faster than the main competitors at detecting actions at test time (\S~\ref{subsec:discussion}).
% Fabio: add sentence on relative test speed comparison?
Thanks to our two-pass action tube generation algorithm, 
in contrast to most existing action classification~\cite{wang-2011,wang-2013,Shuiwang-2013,Karpathy-2014,Simonyan-2014} 
and localisation~\cite{Georgia-2015a, Weinzaepfel-2015} approaches,
our method is capable of detecting and localising multiple co-occurring
action instances in temporally untrimmed videos (see Fig.~\ref{fig:introductionTeaser}).

\section{Related work}

\iffalse
%----------------action classification---------------------
Many approaches to action recognition \cite{laptev-2008,wang-2011,wang-2013} are based on appearance 
(e.g. HOG \cite{dalal-2005}, SIFT \cite{Lowe-2004}) or motion features 
(e.g., optical flow, MBH \cite{dalal-2006}),
encoded using Bag of Visual Words \cite{csurka2004visual} or Fisher vectors \cite{perronnin2007fisher}
The resulting descriptors are typically used to train classifiers (e.g. SVM) in order to predict the labels of action videos.
\fi

%-----------------cnn based action classification and localisation--------------------
Recently, inspired by the record-breaking performance of CNNs in image classification \cite{krizhevsky2012} 
and object detection from images \cite{girshick-2014},
deep learning architectures have been increasingly applied to action classification \cite{Shuiwang-2013,Karpathy-2014,Simonyan-2014}, spatial \cite{Georgia-2015a} or spatio-temporal 
\cite{Weinzaepfel-2015} action localisation,  
and event detection \cite{xu2014discriminative}. 
\iffalse %Repetition.
Indeed %, inspired by this success %of \cite{Georgia-2015a} and \cite{Simonyan-2014}, 
we also extract CNN features, both from RGB and optical flow-derived motion images,
but instead of concatenating the features and passing them to a classifier,
we perform region proposal and detection simultaneously using \emph{a single Faster-RCNN network, trained end-to-end,
and then merge the results using a score compatibility function}.
\fi
% localisation (spatial or spatio-temporal)

The %\textbf{action localisation} 
action localisation problem, in particular, can be addressed by leveraging video segmentation methods.
An example is the unsupervised greedy agglomerative clustering approach of \cite{jain2014tublet}, which resembles Selective Search space-time video blocks.
Since \cite{jain2014tublet} does not exploit the representative power of CNN features, they fail to achieve state-of-the-art results.
Soomro~\etal~\cite{Soomro2015} learn the contextual 
relations between different space-time video segments. Such `supervoxels', however, may end up spanning very long time intervals, failing to localise each action instance individually.
Similarly, \cite{vanGemert2015apt} uses unsupervised clustering to generate a small set of bounding box-like spatio-temporal action proposals.
%Although the authors rely on the branch-and-bound search algorithm \cite{yuan2011discriminative}, %with the addition of motion trajectories, 
However, since the approach in \cite{vanGemert2015apt} employs dense-trajectory features 
%which rely on motion 
\cite{wang-2011},
it does not work on actions characterised by small motions \cite{vanGemert2015apt}.

% context (also to improve localisation)
\iffalse
\textbf{Contextual cues} can be used to improve on action recognition and localisation. 
Whereas \cite{Georgia-2015b} is limited to localisation in still images,
\cite{Jain-2015} will only reap benefits where the objects involved in the action are discriminative.
By contrast, \emph{our method is data-driven and can discriminate between distinct actions categories involving the same objects}
(e.g. the `door' in LIRIS-HARL's `unsuccessfully unlocking door' and `successfully unlocking door' classes).
\fi

% temporal localisation
The %\textbf{temporal detection}
temporal detection of actions~\cite{jiang2014thumos,gorban2015thumos} 
and gestures~\cite{escalera2014chalearn} in temporally untrimmed videos has also recently attracted much interest \cite{yeung2015every, Evangel-2014}.
%These challenges led to big advances in the state-of-the-art \cite{yeung2015every}.
Sliding window approaches have been extensively used \cite{laptev2007retrieving,gaidon2013temporal,tian2013spatiotemporal,wang2014video}. 
%On the contrary, Evangelidis \etal \cite{Evangel-2014} show efficiency for dynamic programming for temporal trimming. 
%For instance, Tian et al. \cite{tian2013spatiotemporal} extend deformable part models to 3D and employed a sliding window which varied in scale, space and time.
%Wang \etal  \cite{wang2014video} first used a temporal sliding window and then modelled the relations between dynamic-poselets. 
Unlike our approach, these methods \cite{tian2013spatiotemporal,wang2014video,yeung2015every} only address \emph{temporal} detection,
and suffer from the inefficient nature of temporal sliding windows.
Our framework is based on incrementally linking frame-level region proposals and temporal smoothing (in a similar fashion to \cite{Evangel-2014}), 
{an approach which is computationally more efficient and can handle long untrimmed videos}.

% spatio-temporal localisation via linking region proposals
Indeed %\textbf{methods which connect frame-level region proposals} 
methods which connect frame-level region proposals for joint spatial and temporal localisation
%(for instance using a human detector to generate bounding box region proposals \cite{yu2015fast})
have risen to the forefront of current research.
% our main competitors
%As clearly stated in the Introduction,
%we identify \cite{Georgia-2015a} and \cite{Weinzaepfel-2015} as our main competitors.
Gkioxari and Malik \cite{Georgia-2015a} have extended \cite{girshick-2014} 
and \cite{Simonyan-2014} to tackle action detection
%simultaneous action classification and localisation,
using unsupervised Selective-Search region proposals and separately trained SVMs.
However, as the videos used to evaluate their work only contain one action and were already temporally trimmed (J-HMDB-21~\cite{J-HMDB-Jhuang-2013}),
it is not possible to assess their temporal localisation performance.  
Weinzaepfel \emph{et al.}'s approach \cite{Weinzaepfel-2015}, instead, first generates region proposals using EdgeBoxes \cite{zitnick2014edge} at frame level to later 
use a tracking-by-detection approach based on a novel track-level descriptor called a Spatio-Temporal Motion Histogram. % Fabio: to do what? construct action paths/tubes?
Moreover, \cite{Weinzaepfel-2015} achieves temporal trimming using a multi-scale sliding window over each track, making it inefficient for longer video sequences.
Our approach improves on both \cite{Georgia-2015a, Weinzaepfel-2015} by using an efficient two-stage single network for detection of region proposals  
and two passes of dynamic programming for tube construction.

\iffalse
In opposition \emph{our approach is more efficient, as we use no intermediate representation and solve the linking problem efficiently via dynamic programming}.
This is shown in the results of Table \ref{table:speed}.
\fi

% include this paper in SOA- End-to-end Learning of Action Detection from Frame Glimpses in Videos

%\textbf{Multiple action detection}.
Some of the reviewed approaches \cite{Weinzaepfel-2015,vanGemert2015apt} could potentially be able to detect co-occurring actions.
However, \cite{Weinzaepfel-2015} {limit their method to produce maximum of two detections per class}, 
while \cite{vanGemert2015apt} does so on the MSRII dataset \cite{MSRIIcao2010cross} 
which only contains three action classes of repetitive nature (clapping, boxing and waving).
Klaser at al. \cite{klaser-2010} use a space-time descriptor and a sliding window classifier to detect the location of only two actions (phoning and standing up).
{In contrast, in our LIRIS-HARL tests (\S~\ref{sec:results_liris}) we consider 10 diverse action categories}.

\iffalse
In \cite{yeung2015every} a Long Short Term Memory network is proposed  
for multiple co-occurring action detection, %Fabio: sounds like we should say more on this paper
but spatial localisation is ignored.
\fi

%-  MAYBE A TABLE COMPARING OUR FEATURES TO THOSE OF MAIN COMPETITORS COULD BE GOOD TO PUT

\section{Methodology} \label{sec:methodology}

As outlined in Figure \ref{fig:algorithmOverview}, our approach combines a region-proposal network (\S~\ref{subsec:rpn}-Fig.~\ref{fig:algorithmOverview}b) with
a detection network (\S~\ref{subsec:frcnn}-Fig.~\ref{fig:algorithmOverview}d),
and fuses the outputs (\S~\ref{subsec:fs_spatil_flow}-Fig.~\ref{fig:algorithmOverview}f)
to generate action tubes (\S~\ref{subsec:action_tube}-Fig.~\ref{fig:algorithmOverview}g-h).
All components are described in detail below.

\iffalse
% this figure is nice, but probably not necessary given the good description in the text
\begin{figure*}[ht!]
  \centering
  \includegraphics[scale=0.23]{figures/methodology/rpn.pdf}
  \vspace{-6mm}
  \caption{Region Proposal Network architecture~\cite{ren2015faster}.}
  \vspace{-6mm}
  \label{fig:rpn}
\end{figure*}
\fi

%\subsection{Network architectures for box regression and classification}\label{sec:network-architectures}

\subsection{Region Proposal Network}\label{subsec:rpn}

To generate rectangular action region hypotheses in a video frame we adopt the Region Proposal Network (RPN) approach of~\cite{ren2015faster}, %(Fig.~\ref{fig:rpn}), 
%which uses the very deep
which is built on top of the last convolutional layer of the 
VGG-16 architecture by Simonyan and Zisserman~\cite{simonyan2014very}. 
% --------------------------------- details on RPN architecture -----------------------------------------
%A RPN is modelled as a small network built on the top of the last convolutional layer
%(the 13-th layer named \emph{conv5\_3})
%of a VGG-16 net.
To generate region proposals, this mini-network slides over the convolutional feature map outputted
by the last layer,
%(\emph{conv5\_3}),
processing at each location an $n\times n$ spatial window and mapping it to a lower dimensional feature vector (512-d for VGG-16).
The feature vector is then passed to two fully connected layers:
a box-regression layer
and a box-classification layer.

During training, for each image location,
$k$ region proposals (also called `anchors') \cite{ren2015faster} are generated.
%TODO put this detail in the supplementary material.
%\footnote{We use $k=9$, considering $3$ scales and $3$ aspect ratios, at each sliding position.}. 
We consider those anchors with a high Intersection-over-Union ($IoU$) with the ground-truth boxes ($IoU > 0.7$) as positive examples,
whilst those with $IoU < 0.3$ as negatives.
% criterion unclear - the highest even when IoU<=0.7? or a combination of the two?
% and the rest as negatives - this statement is worng - the anchors which has IoU between 0.3 to 0.7 don't contribute to the training
Based on these training examples, the network's objective function is minimised using stochastic gradient descent (SGD),
%(please refer to~\cite{ren2015faster} for more details)
encouraging the prediction of both 
the probability of an anchor belonging to action or no-action category (a binary classification),
and the 4~coordinates of the bounding box.
%(i.e. regression of an anchor box to its nearby ground-truth box).
%The predicted box is a regressed box from an anchor to a nearby ground-truth box.
%
%using non-maximal suppression (NMS). %(after Non-Maximal Suppression, NMS) per frame.  
%At testing time we retain the top 300 (after NMS) region proposals per frame, based on their actionness scores.

\subsection{Detection network} \label{subsec:frcnn}

For the detection network we use a Fast R-CNN net~\cite{girshick2015fast} with a VGG-16 architecture~\cite{simonyan2014very}. 
This takes the RPN-based region proposals (\S~\ref{subsec:rpn}) 
and regresses a new set of bounding boxes for each action class and associates classification scores.
Each RPN-generated region proposal leads to $C$ (number of classes) regressed bounding boxes with corresponding class scores.

Analogously to the RPN component, the detection network is also built upon the convolutional feature map outputted by the last layer of the VGG-16 network.
It generates a feature vector for each proposal generated by RPN,
which is again fed to two sibling fully-connected layers:
a box-regression layer
and a box-classification layer.
%1) a classification layer 2) a box-regression layer. 
Unlike what happens in RPNs, these layers produce $C$ multi-class softmax scores and refined boxes (one for each action category) for each input region proposal.
%%The network first processes the entire video frame using 13 convolutional and 4 max pooling layers to generate a feature map.
%The resulting feature maps output from the network are processed by a sequence of fully connected layers and linked to two final output stages:
%1) a classification layer, which generates a softmax probability estimate for one of the action classes per region proposal,
%2) a box-regression layer, which generates a regressed box for each input region proposal \cite{girshick2015fast}.
%\subsection{CNN training strategy}\label{subsec:training_rpn_frcnn}
\vspace{-3mm}
\paragraph{CNN training strategy.}\label{subsec:training_rpn_frcnn}
We employ a variation on the training strategy of~\cite{ren2015faster} to train both the RPN and Fast R-CNN networks.
Shaoqing \etal\ ~\cite{ren2015faster} suggested a 4-steps `alternating training' algorithm in which in the first 2~steps,
a RPN and a Fast R-CNN nets are trained independently, while in the 3$^{\textnormal{rd}}$ and 4$^{\textnormal{th}}$ steps the two networks are fine-tuned
with shared convolutional layers.
%--- REVIEWER-1:  it is claimed that “in practice we found that the detection accuracy decreases with shared convolutional features”. Provide a brief explanation. ---
%\textcolor{red}{
In practice, we found empirically that the detection accuracy on UCF101 slightly decreases when using shared convolutional features, i.e.,
when fine tuning the RPN and Fast-RCNN trained models obtained after the first two steps.
As a result, we train the RPN and the Fast R-CNN networks independently following only the 1$^{\textnormal{st}}$ and 2$^{\textnormal{nd}}$ steps of~\cite{ren2015faster},
while neglecting the 3$^{\textnormal{rd}}$ and 4$^{\textnormal{th}}$ steps suggested by~\cite{ren2015faster}.
%}
\subsection{Fusion of appearance and motion cues}\label{subsec:fs_spatil_flow}

In a work by Redmon \etal\ \cite{redmon2015you},
the authors combine the outputs from Fast R-CNN and YOLO (You Only Look Once) object detection networks to reduce background detections and improve the overall detection quality.
Inspired by their work, we use our motion-based detection network to improve the scores of the appearance-based detection net (c.f.~Fig.~\ref{fig:algorithmOverview}\textbf{f}).

Let $\{\Vector{b}^{s}_i\}$ and $\{\Vector{b}^{f}_j\}$ denote the sets of detection boxes generated by the appearance- and motion-based detection networks, respectively, on a given test frame and
for a specific action class $c$.
%We compute the IoU to quantify the region overlap between each pair of spatial $\Vector{b}^{s}_{i}$ and flow $\Vector{b}^{f}_{j}$ detection boxes.
%For a spatial box $\Vector{b}^{s}_{i}$, we find the best overlap
%by computing the maximum overlap $\delta_{s}$ with the flow boxes.
Let $\Vector{b}^{f}_{max}$ be the motion-based detection box with maximum overlap with a given appearance-based detection box $\Vector{b}^{s}_{i}$. 
If this maximum overlap, quantified using the IoU,
is above a given threshold $\Scalar{\tau}$,
we augment the softmax score $s_{c}(\Vector{b}^{s}_{i})$ of the appearance-based box %with the following compatibility function: % for all action classes $c$ 
as follows: 
%We extract equal number of spatial and flow detection boxes and thus $i,j \in \{ 1,\dot,\Scalar{D}\}$.
\begin{equation}
\label{eqn:appflow_fusion}
s_{c}^{*}(\Vector{b}^{s}_{i}) = s_{c}(\Vector{b}^{s}_{i}) + s_{c}(\Vector{b}^{f}_{max})\times IoU(\Vector{b}^{s}_{i}, \Vector{b}^{f}_{max}).
%- \abs{s_{c}(\Vector{b}^{s}_{i}) - s_{c}(\Vector{b}^{f}_{max})}.
\end{equation} % Fabio: why did you strike out the last term?
The second term adds to the existing score of the appearance-based detection box a proportion, equal to the amount of overlap, of the motion-based detection score.
%, while the third term penalises it as per the difference of their class-specific probability scores.
%TODO put this back if it's implemented: Finally, we normalise the augmented scores $s_{c}^{*}(\Vector{b}^{s}_{i})$ of all spatial detection boxes.
In our tests we set $\Scalar{\tau}=0.3$.

\subsection{Action tube generation} \label{subsec:action_tube}
\label{sec:dynamicProgramming}

The output of our fusion stage (\S~\ref{subsec:fs_spatil_flow}) is, for each video frame,
a collection of detection boxes for each action category,
together with their associated augmented classification scores~(\ref{eqn:appflow_fusion}).
%We now turn to the task of associating the 
Detection boxes can then be linked up in time to identify video regions most likely to be associated with a single action instance, or \emph{action tube}.
%we are in a position to identify sequences of detection boxes most likely to be associated with a single action instance, which we call \emph{action tubes}.
Action tubes are connected sequences of detection boxes in time, without interruptions, 
%such that their classification score is high for the considered class, and consecutive detections show significant spatial overlap. Note that, 
and unlike those in~\cite{Georgia-2015a} they
are not constrained to span the entire video duration.
%We obtain temporal detection 
%This is achieved by a second pass of dynamic programming which takes care of temporal detection.

They are obtained as solutions to two consecutive energy maximisation problems.
%In the first pass we construct 
First a number of action-specific paths $\Vector{p}_{c}=\{ \Vector{b}_{\Index{1}}, \dots,  \Vector{b}_{T} \}$,
spanning the entire video length, are constructed by linking detection boxes over time in virtue of their class-specific scores
and their temporal overlap.
Second, action paths are temporally trimmed by ensuring that the constituting boxes' detection scores are consistent with the foreground label $c$.
%Both maximisation problems are solved by dynamic programming.
\vspace{-3mm}
\paragraph{Building action paths.} \label{sec:building-paths}
We define the energy $E(\Vector{p}_c)$ for a particular path $\Vector{p}_c$ linking up detection boxes for class $c$ across time to be the a sum of unary and pairwise potentials:
\begin{equation}
  \label{eqn:firstpassenergy}
  E(\Vector{p}_c) = \sum_{t=1}^T s^*_c(\Vector{b}_t) + \lambda_{o} \sum_{t=2}^T \psi_{o} \left( \Vector{b}_t, \Vector{b}_{t-1} \right),
\end{equation}
where $s^*_c(\Vector{b}_t)$ denotes the augmented score (\ref{eqn:appflow_fusion}) of detection $\Vector{b}_t$,
the overlap potential $\psi_{o}(\Vector{b}_{\Index{t}}, \Vector{b}_{\Index{t}-1})$ is the IoU of the two boxes $\Vector{b}_t$ and $\Vector{b}_{t-1}$,
and $\lambda_{o}$ is a scalar parameter weighting the relative importance of the pairwise term.
The value of the energy (\ref{eqn:firstpassenergy}) is high for paths whose detection boxes score highly for the particular action category $c$, and for which
consecutive detection boxes overlap significantly.
We can find the path which maximises the energy,
$
  \label{eq:region_max}
  \Vector{p}_{c}^* = \textnormal{argmax}_{\Vector{p}_{c}} \ E(\Vector{p}_c)
$, by simply applying the Viterbi algorithm \cite{Georgia-2015a}.

Once an optimal path has been found, we remove all the detection boxes associated with it and recursively seek the next best action path.
%until no more can be found series of detections spanning the whole video can be found. 
% For sake of computational efficiency we extract up to three best paths - 
%Fabio: does this mean we are actually constrained to detecting a maximum of 3 co-occurring instances?
Extracting multiple paths allows our algorithm to account for multiple co-occurring instances of the same action class.
\vspace{-3mm}
\paragraph{Smooth path labelling and temporal trimming.} \label{sec:temporal-trimming}

As the resulting action-specific paths span the entire video duration, while human actions typically only occupy a fraction of it, temporal trimming becomes necessary.
The first pass of dynamic programming~(\ref{eqn:firstpassenergy}) aims at extracting connected paths by penalising regions which do not overlap in time.
As a result, however, not all detection boxes within a path exhibit strong action-class scores.
%We now aim to trim the paths and form action tubes which exhibit consecutively high-scoring detection boxes.
%However, since paths are extracted for each action category independently, 
%the final labels assigned to each region proposal in time need not be consistent or smooth.

The goal here is to assign to every box $\Vector{b}_t \in \Vector{p}_{c}$ in an action path $\Vector{p}_{c}$ a binary label \mbox{${l}_{t}$ $\in$ $\{c,0\}$}
(where zero represents the `background' or `no-action' class),
subject to the conditions that  the path's labelling ${\Vector{L}_{\Vector{p}_{c}}} = [ l_1, l_2, \dots, l_T]'$:
i) is consistent with the unary scores~(\ref{eqn:appflow_fusion});
and ii) is smooth (no sudden jumps).\\
As in the previous pass, we may solve for the best labelling by maximising:
\begin{equation}
  \label{eqn:secondpassenergy}
  \Vector{L}_{\Vector{p}_{c}}^* = \underset{\Vector{L_{\Vector{p}_{c}}}}{\textnormal{argmax}} \ \left( \sum_{t=1}^T s_{l_t}(\Vector{b}_t) - \lambda_{l} \sum_{t=2}^T \psi_l \left( \Scalar{l}_{t}, \Scalar{l}_{t-1} \right) \right),
\end{equation}
%where $\Vector{L}_p~=~(l_1,l_2,l_3, \dots l_T)$, is a sequence of $0$ and $1$ labels for a path $p$, and 
where 
$\lambda_{l}$ is a scalar parameter weighting the relative importance of the pairwise term.
%$s_{l_t=1}(\Vector{b}_t) = s^*_c(\Vector{b}_t)$, $s_{l_t=0}(\Vector{b}_t) = 1 - s^*_c(\Vector{b}_t)$, and $s^*_c(\Vector{b}_t)$ are the augmented scores (\ref{eqn:appflow_fusion}).\\
The pairwise potential $\psi_l$ is defined to be:
\vskip -0.5cm
\begin{equation}
  \label{eqn:potts}
  \psi_l(l_t,l_{t-1}) = \begin{cases} 0 &\mbox{if } l_t=l_{t-1}\\
  \alpha_c & \mbox{otherwise} , \end{cases}
\end{equation}
where $\alpha_c$ is a class-specific constant parameter which we set by cross validation.
In the supplementary material, we show the impact of the class-specific $\alpha_c$ on the detection accuracy.
Equation (\ref{eqn:potts}) is the standard Potts model which penalises labellings that are not smooth, thus enforcing a piecewise constant solution.
Again, we solve (\ref{eqn:secondpassenergy}) using the Viterbi algorithm. 
 
All contiguous subsequences of the retained action paths $\Vector{p}_c$ associated with category label $c$ constitute our action tubes. 
As a result, one or more distinct action tubes spanning arbitrary temporal intervals may be found in each video for each action class $c$. % Fabio: say in introduction too
%For each action tube so extracted from $\Vector{R}_{c}$, with initial frame $t_{s}$ and final frame $t_{e}$, we have a vector of scores $\Vector{S} = ( s_c(\Vector{r}_{t_s}),s_c(\Vector{r}_{t_{s+1}}), \dots ,s_c(\Vector{r}_{t_e}) )$ for class~$c$.
Finally, 
%each action tube is assigned a global score equal to the mean of the top $k$ box detection scores within it.
each action tube is assigned a global score equal to the mean of the top $k$ augmented class scores~(\ref{eqn:appflow_fusion}) of its constituting detection boxes. 

\section{Experimental validation and discussion} \label{subsec:discussion}

%\subsection{Datasets and performance measures} \label{datasets:sec:detection}

In order to evaluate our spatio-temporal action detection pipeline we selected what are currently considered among the most challenging action detection datasets:
UCF-101~\cite{soomro-2012}, LIRIS HARL D2~\cite{liris-harl-2012}, and J-HMDB-21~\cite{J-HMDB-Jhuang-2013}.
%We selected 
%\noindent
UCF-101 is the largest,
most diverse and challenging dataset to date,
and contains realistic sequences with a large variation in camera motion,
appearance, human pose, scale, viewpoint,
clutter and illumination conditions.
Although each video only contains a single action category, it may contain multiple action instances of the same action class.
To achieve a broader comparison with the state-of-the-art, we also ran tests on the J-HMDB-21~\cite{J-HMDB-Jhuang-2013} dataset.
The latter is a subset of HMDB-51~\cite{HMDBkuehne2011hmdb} with 21 action categories and 928 videos, each 
containing a single action instance and trimmed to the action's duration.
The reported results were averaged over the three splits of J-HMDB-21.
Finally we conducted experiments on the more challenging LIRIS-HARL dataset, which contains 10 action categories,
including human-human interactions and human-object interactions 
(e.g., `discussion of two or several people', and `a person types on a keyboard'\footnote{http://liris.cnrs.fr/voir/activities-dataset}).
In addition to containing multiple space-time actions,
some of which occurring concurrently,
the dataset contains scenes where relevant human actions take place amidst other irrelevant human motion.
%On LIRIS-HARL, we evaluate the efficacy of our proposed 2-pass DP algorithm, by comparing it to a one-pass method (see Table~\ref{table:liris_results_1}).

For all datasets we used the exact same evaluation metrics and data splits as in the original papers.
In the supplementary material, we further discuss all implementation details, and propose an interesting % together with a
quantitative comparison between Selective Search- and RPN-generated region proposals.
 
\paragraph{Performance comparison on UCF-101.}
Table~\ref{table:ucf101_results_1} presents the results we obtained on UCF-101, and 
compares them to the previous state-of-the-art~\cite{Weinzaepfel-2015,yu2015fast}.
We achieve an mAP of $66.75\%$ compared to $46.77\%$ reported by~\cite{Weinzaepfel-2015} (a $20\%$ gain), at the standard threshold of $\delta~=~0.2$.
At a threshold of $\delta=0.4$ we still get a high score of $46.35\%$,
(comparable to $46.77\%$~\cite{Weinzaepfel-2015} at $\delta=0.2$).
\iffalse
% TODO: Move to supp 
In order to obtain a fair comparison with \cite{Georgia-2015a}, we implemented their multi-stage pipeline and added a second pass of DP for temporal localisation
(in addition to the one pass in~\cite{Georgia-2015a}). 
In Table~\ref{table:ucf101_results_1}, this method is denoted as `ActionTube'. 
We found that the recall of Selective Search detection boxes degrades as the IoU threshold increases - one reason why `ActionTube' performs poorly.
\fi
Note that we are the first to report results on UCF-101 up to $\delta=.6$,
attesting to the robustness of our approach to more accurate localisation requirements.
Although our separate appearance- and motion-based detection pipelines already outperform the state-of-the-art (Table~\ref{table:ucf101_results_1}), 
their combination (\S~\ref{subsec:fs_spatil_flow}) delivers a significant performance increase.
\begin{table}[tb]
  \centering
  \caption{Quantitative action detection results (mAP) on the UCF-101 dataset.}
  {\footnotesize
  \scalebox{0.9}{
  \begin{tabular}{lccccccc}
  \toprule
  %\hline
  Spatio-temporal overlap threshold $\delta$ & 0.05 & 0.1 & 0.2 & 0.3 & 0.4 & 0.5 & 0.6\\ \midrule
  %\hline\hline
  FAP~\cite{yu2015fast} & 42.80 & -- & -- & -- & -- & -- & --\\
  %\hline
  % TODO: Move to supp ActionTube~\cite{Georgia-2015a} & -- & 43.0 & -- & -- & -- & -- & --\\
  %\hline
  STMH~\cite{Weinzaepfel-2015} & 54.28 & 51.68 & 46.77 & 37.82 & -- & -- & --\\ \midrule
  %\hline
  Our (appearance detection model) & 68.74  &  66.13 &  56.91 &  48.28  & 39.10 &  30.67 &  22.77 \\                    
  %\hline
  Our (motion detection model)    & 67.04  &  64.86 &  57.33  & 47.45  & 38.65 &  28.90 & 19.49 \\                   
  %\hline
  \textbf{Our (appearance $+$ motion fusion)} & \textbf{79.12} &  \textbf{76.57} &  \textbf{66.75} &  \textbf{55.46} &  \textbf{46.35} &  \textbf{35.86} &  \textbf{26.79} \\
  %\hline
  \bottomrule 
  \end{tabular}
  }
  }
  \vspace*{-\baselineskip}
  \label{table:ucf101_results_1}
\end{table} 

%--- added by suman on May 4 10:12 pm ----
%Fabio: would be nice to provide more evidence of our multiple instance detection capabilities, e.g. as basketball sequence?

\begin{figure}[b]
  \centering
  \includegraphics[width=0.98\textwidth]{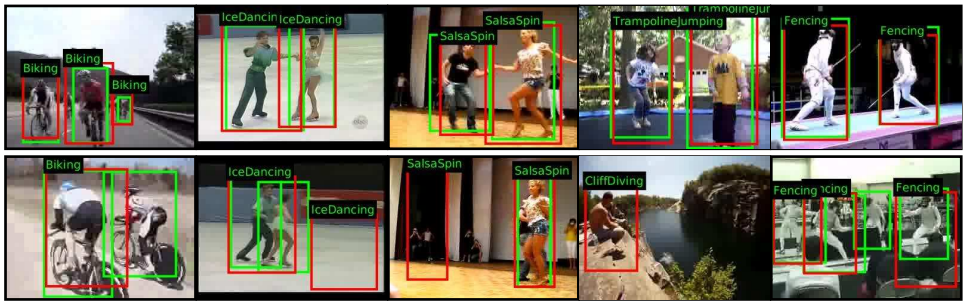}
  \caption{
    {\small
      \textit{
        Action detection/localisation results on UCF101.
        Ground-truth boxes are in green, detection boxes in red.
        The top row shows correct detections,
        the bottom one contains examples of more mixed results. 
        In the last frame, 3 out of 4 `Fencing' instances are nevertheless correctly detected.
      }
    }
  }
  \label{fig:ucf101_res_1}
\end{figure}

Some representative example results from UCF-101 are shown in Fig.~\ref{fig:ucf101_res_1}.
%Unlike~\cite{Weinzaepfel-2015}, 
Our method can detect several (more than $2$) action instances concurrently,
as shown in Fig.~\ref{fig:introductionTeaser}, in which three concurrent instances and in total six action instances are detected correctly.
%and in the first frame of Fig.~\ref{fig:ucf101_res_1}  where three `biking' instances in one particular frame are successfully isolated.
%--- REVIEWER-1: The article claimed that the proposed method can detect and localise the concurrent actions in videos.
% This is not validated properly as there are only six concurrent instances 
% in the UCF101 dataset. Authors should have considered UT-interaction dataset.
%\textcolor{red}{
Quantitatively, we report class-specific video AP (average precision in $\%$) of $88.0$, $83.0$ and $62.5$ on the UCF-101 action categories \emph{`Fencing'}, \emph{`SalsaSpin'} and 
\emph{`IceDancing'}, 
respectively, which all concern multiple inherently co-occurring action instances.
%} 
Class-specific video APs on UCF-101 are reported in the supplementary material. %for `{spatial}', `{flow}' and `{fused}' detection pipelines. 

\iffalse
where our method successfully detect three instances of the `Biking' class in both space and time.
\fi

% Spatio temporal detection results on UCF-101 and spatial localisation results on J-HMDB-21.

% \paragraph{Impact of the $2^{nd}$ pass of DP}
% To demonstrate the efficacy of our $2^{nd}$ pass of \emph{Viterbi} algorithm, we generate results using the \emph{spatial+flow$*$ fusion}
% (\S~Table \ref{table:ucf101_results_1}) detection model by setting the alpha $\alpha=0$ for all the classes.
% With this setting, the action paths are trimmed only based on the frame-level detection 
% scores without putting any constraint ($\alpha$) on the label consistency.
% We can observe a performance decrease (a $6\%$ decreases in mAP at $\delta=20\%$) with this setting.
% This shows that our $2^{nd}$ pass of DP helps in label smoothing when there present any sudden flickering in the detection scores.

\begin{figure}[ht!]
  \centering
  \includegraphics[width=0.98\textwidth]{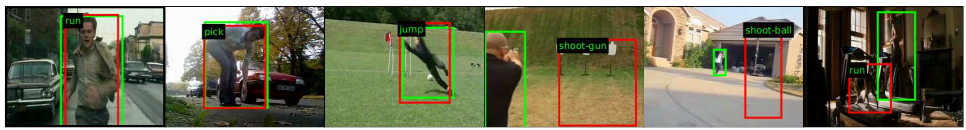}
  \caption{
    {\small
      \textit{
        Sample space-time action localisation results on JHMDB. Left-most three frames: accurate detection examples. Right-most three frames: mis-detection examples.
      }
    }
  }
  \label{fig:J-HMDB_results}
\end{figure}

\paragraph{Performance comparison on J-HMDB-21.}\label{sec:results_jhmdb}
The results we obtained on J-HMDB-21 are presented in Table~\ref{table:jhmdb21_results_1}.
Our method again outperforms the state-of-the-art, with an mAP increase of $18\%$ and $11\%$ at $\delta=.5$ as compared to \cite{Georgia-2015a} and \cite{Weinzaepfel-2015}, respectively. 
Note that our motion-based detection pipeline alone exhibits superior results,
and when combined with appearance-based detections leads to a further improvement of $4\%$ at $\delta=.5$.
%\paragraph{High precision of detections}
%Note also that our detection results in both Table~\ref{table:ucf101_results_1} (UCF-101) and~\ref{table:jhmdb21_results_1} (J-HMDB-21) are robust to increasing threshold values $\delta$.
These results attest to the high precision of the detections -
a large portion of the detection boxes have high IoU overlap with the ground-truth boxes,
a feature due to the superior quality of RPN-based region proposals as opposed to Selective Search's
(a direct comparison is provided in the supplementary material). %(compare to `Action Tube' which is based on Selective-Search').
Sample detections on J-HMDB-21 are shown in Figure \ref{fig:J-HMDB_results}.
Also, we list our classification accuracy results on J-HMDB-21 in Table~\ref{table:jhmdb21_results_acc},
where it can be seen that our method achieves an $8\%$ gain compared to~\cite{Georgia-2015a}.
%Our method classifies J-HMDB-21 action categories more accurately ($8\%$ better) than~\cite{Georgia-2015a}.

% Fabio: shall we mention that we have proof of that in the additional material?

\begin{table}[ht!]
\centering
\footnotesize
\caption{Quantitative action detection results (mAP) on the J-HMDB-21 dataset.}
%\begin{center}
\scalebox{0.9}{
\begin{tabular}{lcccccccc}
%\hline
\toprule
Spatio-temporal overlap threshold $\delta$  & 0.1 & 0.2 & 0.3 & 0.4 & 0.5 & 0.6 & 0.7 \\ \midrule
%\hline\hline
ActionTube~\cite{Georgia-2015a} (mAP) & -- & -- & -- & -- & 53.3 & -- & --\\
Wang~\etal~\cite{WangQTV16} (mAP) & -- & -- & -- & -- & 56.4 & -- & --\\
%\hline
STMH~\cite{Weinzaepfel-2015} (mAP) & -- & 63.1 & 63.5 & 62.2 & 60.7 & -- & -- \\ \midrule
%\hline
Our (appearance detection model) (mAP) & 52.99 & 52.94 & 52.57 & 52.22 & 51.34 & 49.55 & 45.65 \\
%\hline
Our (motion detection model) (mAP)  & 69.63 & 69.59 & 69.49 & 69.00 & 67.90 & 65.25 & 54.35  \\
%\hline
\textbf{Our (appearance+motion fusion)} (mAP) & \textbf{72.65} & \textbf{72.63} & \textbf{72.59} & \textbf{72.24} & \textbf{71.50} & \textbf{68.73} & \textbf{56.57} \\
%\hline
\bottomrule 
\end{tabular}
}
%\end{center} \vspace{-6mm}
\vspace*{-\baselineskip}
\label{table:jhmdb21_results_1}
\end{table} 

\begin{table}[ht!]
\centering
\footnotesize
\caption{Classification accuracy on the J-HMDB-21 dataset.}
%\begin{center}
\scalebox{0.9}{
\begin{tabular}{l|cccc}
%\hline
\toprule
Method & Wang \etal~\cite{wang-2011} & STMH~\cite{Weinzaepfel-2015}  & ActionTube~\cite{Georgia-2015a} & Our (appearance+motion fusion) \\ \midrule
\textbf{Accuracy (\%)} & 56.6 & 61 & 62.5 & \textbf{70.0} \\
\bottomrule 
\end{tabular}
}
%\end{center} \vspace{-6mm}
\vspace*{-\baselineskip}
\label{table:jhmdb21_results_acc}
\end{table} 

\paragraph{Performance comparison on LIRIS-HARL.}\label{sec:results_liris}
LIRIS HARL allows us to demonstrate the efficacy of our approach on temporally un-trimmed videos with co-occurring actions.
For this purpose we use LIRIS-HARL's specific evaluation tool - the results are shown in Table~\ref{table:perf_integrated}.
Our results are compared with those of i) VPULABUAM-13~\cite{liris-teamp-13} and ii) IACAS-51~\cite{liris-teamp-51} 
from the original LIRIS HARL detection challenge.
In this case, our method outperforms the competitors by an even {larger} margin.
We report space-time detection results
by fixing the threshold quality level to 10\% for the four thresholds~\cite{liris-harl-2014}
and measuring temporal precision and recall along with spatial precision and recall, to produce an integrated score.
We refer the readers to~\cite{liris-harl-2014} for more details on LIRIS HARL's evaluation metrics.
\begin{figure}[ht!]
  \centering
  \includegraphics[width=0.98\textwidth]{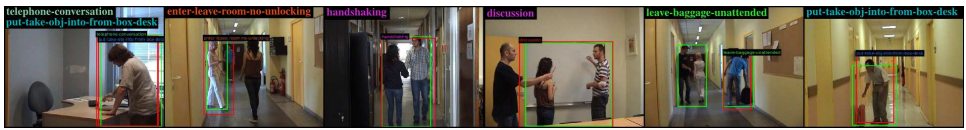}
  \caption{
    {\small
      \textit{
        Frames from the space-time action detection results on LIRIS-HARL,
        some of which include single actions involving more than one person like `handshaking' and `discussion'.
        Left-most three frames: accurate detection examples. Right-most three frames: mis-detection examples.
        %More examples are shown in the attached supplementary video.
%         in the 4-th and 6-th frames the predicted labels are ``discussion'' (ground-truth: ``give-object-to-person'') and  
% 	``put-take-obj-into-from-box-desk'' (ground-truth: ``leave-baggage-unattended'') respectively;
% 	whereas in 5-th frame one action instance (``leave-baggage-unattended'') is correctly detected and
% 	another instance (``handshaking'') is not detected at all.
      }
    }
  }
  \label{fig:liris_results}
  \vskip -0.4cm
\end{figure}

We also report in Table~\ref{table:liris_results_1} 
the mAP scores obtained by the appearance, motion and the fusion detection models, respectively (note that there is no prior state of the art to report in this case).
Again, we can observe an improvement of $7\%$ mAP at $\delta=.2$ due to our fusion strategy. 
To demonstrate the advantage of our 2nd pass of DP (\S~\ref{sec:temporal-trimming}), we also generate results (mAP) using only the first DP pass (\S~\ref{sec:building-paths}).
\iffalse
, we generate 
class specific action paths for each video which are temporally untrimmed and report the mAP at $\delta=20\%$.
\fi
%We denote this one-pass result in Table \ref{table:liris_results_1} as \emph{spatial+flow fusion (*)}.
Without the 2$^\textnormal{nd}$ pass performance decreases by $20\%$, highlighting the 
importance of temporal trimming in the construction of action tubes.
%need for temporal trimming with a second pass of DP for space-time action detection in temporally un-trimmed videos.
\begin{table}[!ht]
\centering
\footnotesize
\caption{Quantitative action detection results on the LIRIS-HARL dataset.}
%\begin{center}
\scalebox{0.9}{
\begin{tabular}{lcccccccc}
%\hline 
\toprule
Method & Recall\tiny{-10} & Precision\tiny{-10} & F1-Score\tiny{-10} & $I_{sr}$ & $I_{sp}$ & $I_{tr}$ &  $I_{tp}$ & IQ \\ \midrule
%\hline\hline
VPULABUAM-13-IQ~\cite{liris-teamp-13} & 0.04 & 0.08 & 0.05 & 0.02 & 0.03 & 0.03 & 0.03 & 0.03\\
IACAS-51-IQ~\cite{liris-teamp-51} & 0.03 & 0.04 & 0.03 & 0.01 & 0.01 & 0.03 & 00.0 & 0.02\\ \midrule
(\textbf{Ours}) & \textbf{0.568} & \textbf{0.595} & \textbf{0.581} & \textbf{0.5383} & \textbf{0.3402} & \textbf{0.4802} & \textbf{0.4739} & \textbf{0.458} \\
%\hline
\bottomrule
\end{tabular}
}
%\end{center}
\vspace*{-\baselineskip}
\label{table:perf_integrated}
\end{table}
\vspace{-2mm}
\begin{table}[!ht]
\centering
\footnotesize
\caption{Quantitative action detection results (mAP) on LIRIS-HARL for different $\delta$.}
%\begin{center}
\scalebox{0.9}{
\begin{tabular}{lcccccccc}
%\hline
\toprule
Spatio-temporal overlap threshold $\delta$ & 0.1 & 0.2 & 0.3 & 0.4 & 0.5 \\ \midrule
%\hline\hline
Appearance detection model  & 46.21 & 41.94 &  31.38  &  25.22 & 20.43 \\
%\hline
Motion detection model  & 52.76 & 46.58 & 35.54  &  26.11 & 19.28 \\
%\hline
%Spatial+flow fusion without $2^{nd}$ pass of DP  & -- &  29.46 & --  &  -- & -- \\
Appearance+motion fusion with one DP pass  & 38.1 &  29.46 & 23.58  &  14.54 & 9.59 \\
%\hline  
%\textbf{Spatial+flow fusion} with $2^{nd}$ pass of DP  & \textbf{54.18} &  \textbf{49.10} & \textbf{35.91}  &  \textbf{28.03} & \textbf{21.36} \\
\textbf{Appearance+motion fusion} with two DP passes & \textbf{54.18} &  \textbf{49.10} & \textbf{35.91}  &  \textbf{28.03} & \textbf{21.36} \\
%\hline
\bottomrule
\end{tabular}
}
%\end{center} \vspace{-6mm}
\vspace*{-\baselineskip}
\label{table:liris_results_1}
\end{table} 
\vspace{-3mm}
% Fabio: add comparative speed performance increase assessment
\paragraph{Test-time detection speed comparison.}
%\textcolor{red}
Finally, we compared detection speed at test time of the combined region proposal generation and CNN feature extraction approach used in~(\cite{Georgia-2015a,Weinzaepfel-2015})
to our neural-net based, single stage action proposal and classification pipeline on the J-HMDB-21 dataset.% (Table \ref{table:detection_time_analysis}).
We found our method to be $10\times$ faster than~\cite{Georgia-2015a} and $5\times$ faster than~\cite{Weinzaepfel-2015},
%in detecting single frame actions (
with a mean of 113.52 \cite{Georgia-2015a}, 52.23 \cite{Weinzaepfel-2015} and 10.89 (ours) seconds per video, averaged over all the videos in J-HMDB-21 split1. 
More timing comparison details and qualitative results (images and video clips) can be found in the supplementary material.
\vspace{-3mm}
\paragraph{Discussion.}
%--- REVIEWER-1 -  The proposed method performed better than state-of-the-art but the answer/discussion to question “why” is missing. ---
%--- meta REVIEWER-2: A couple reviewers suggest that the authors should provide more insights into what makes the model work, which I totally agree ---
%\textcolor{red}{
% The main reasons why our pipeline shows consistent superior performance
% as compared to the state-of-the-art are as follows.
The superior performance of the proposed method is due to a number of reasons.
1) Instead of using unsupervised region proposal algorithms as in~\cite{uijlings-2013,zitnick2014edge},
our pipeline takes advantage of a supervised RPN-based region proposal approach 
which exhibits better recall values than~\cite{uijlings-2013}~(supplementary-material).
%Unlike~\cite{uijlings-2013,zitnick2014edge}, RPN proposals carry actionness scores which help to rank the proposals.
2) Our fusion technique improves the mAPs (over the individual appearance or motion models) by $9.4\%$, $3.6\%$ and $2.5\%$ 
on the UCF-101, J-HMDB-21 and LIRIS HARL datasets respectively.
We are the first to report an ablation study~(supplementary-material) 
where it is shown that the proposed fusion strategy~(\S~\ref{subsec:fs_spatil_flow})
improves the class-specific video APs of UCF-101 action classes.
%(over both spatial and flow nets) for 18 out of 24 UCF-101 action classes.
3) Our original 2nd pass of DP is responsible for significant improvements in mAP by $20\%$ on LIRIS HARL and $6\%$ on UCF-101~(supplementary-material). 
Additional qualitative results are provided in the supplementary video \footnote{\url{https://www.youtube.com/embed/vBZsTgjhWaQ}},
and on the project web page \footnote{\url{http://sahasuman.bitbucket.org/bmvc2016}}, where the code has also been made available.
%}

%--- REVIEWER-1: The article should provide a discussion on qualitative performance. ---
%--- meta reviewr-1: Reviewers made important observations about what needs to be clarified or re-emphasized in the paper and about the experimental evaluation, stating that it lacks qualitative 
% --- evaluation.
%\textcolor{red}{
% Additional qualitative results can be visualised in the supplementary video at:~\url{https://www.youtube.com/embed/vBZsTgjhWaQ}
% and on the project web page at:~\url{http://sahasuman.bitbucket.org/bmvc2016}.
% We made the MATLAB source code publicly available at~\url{https://bitbucket.org/sahasuman/bmvc2016_code}.
%}
\section{Conclusions and future work} \label{sec:conclusions}

In this paper, we presented a novel human action recognition approach which 
%unlike existing state-of-the-art approaches which typically deal with single action classification and/or localisation problems on temporally trimmed videos,
addresses in a coherent framework the challenges involved in concurrent multiple human action recognition,
spatial localisation and temporal detection, thanks to a novel deep learning strategy for simultaneous detection and classification of region proposals and an improved action tube generation approach.
Our method significantly outperforms the previous state-of-the-art on the most challenging benchmark datasets, for it is capable of handling multiple concurrent action instances and temporally untrimmed videos.
%Most importantly though, we evaluated on LIRIS-HARL D2~\cite{liris-harl-2012} dataset which contains multiple concurrent actions, with instances of the same action class happening at the same time, and where all videos are temporally untrimmed. Our proposed pipeline obtained a score 14.3 times higher than the previous top performer.

Its combination of high accuracy and fast detection speed at test time is very promising for real-time applications, for instance smart car navigation. 
As the next step we plan to make our tube generation and labelling algorithm fully incremental and online, by only using region proposals from independent frames at test time and updating the dynamic programming optimisation step at every incoming frame. 
%In future work we also look into exploring active learning to handle annotation costs.
\vfill

{\small\paragraph{Acknowledgements.} This work was partly supported by ERC grant ERC-2012-AdG 321162-HELIOS, EPSRC grant Seebibyte EP/M013774/1 and EPSRC/MURI grant EP/N019474/1.}

\par\vfill\par

\clearpage

\bibliography{main}
\end{document}